\documentclass[11pt,a4paper]{article}
\usepackage{times,latexsym}
\usepackage{url}
\usepackage[T1]{fontenc}

\usepackage[acceptedWithA]{tacl2018v2}
\usepackage{xspace,mfirstuc,tabulary}

\newcommand{\commentout}[1]{}

\newif\iftaclinstructions
\taclinstructionsfalse % AUTHORS: do NOT set this to true
\iftaclinstructions
\renewcommand{\confidential}{}
\renewcommand{\anonsubtext}{(No author info supplied here, for consistency with
TACL-submission anonymization requirements)}
\newcommand{\instr}
\fi

\iftaclpubformat % this "if" is set by the choice of options

\else

\fi

\usepackage{times}
\usepackage{latexsym}

\usepackage{amsmath}
\usepackage{amsfonts}
\usepackage{booktabs}
\usepackage{graphicx}
\usepackage{xargs}
\usepackage{algorithm}
\usepackage{algpseudocode}
\usepackage{float}
\usepackage{bm}

\usepackage{bbm}
\usepackage{comment}
\usepackage{todonotes}
\usepackage{svg}
\usepackage{mathtools}

\usepackage[english]{babel}
\usepackage{amsthm}
\newtheorem{theorem}{Theorem}

\newtheorem{lemma}[theorem]{Lemma}
\newtheorem{assumption}[theorem]{Assumption}

\newcommand{\R}{\mathbb{R}}

\newcommand{\E}{\mathbb{E}}
\newcommand{\X}{\mathcal{X}}
\newcommand{\Y}{\mathcal{Y}}

\renewcommand{\O}{\mathcal{O}}

\newcommand{\D}{\mathcal{D}}
\newcommand{\Ds}{\mathcal{D}_{s}}

\title{Partially Supervised Named Entity Recognition\\ via the Expected Entity Ratio Loss}

\author{
  Thomas Effland\\
  Columbia University\\
  {\tt teffland@cs.columbia.edu}
  \And
  Michael Collins\\
  Google Research\\
  {\tt mjcollins@google.com}
}

\date{}

\begin{document}
\maketitle
\begin{abstract}
We study learning named entity recognizers in the presence of missing entity annotations.
We approach this setting as tagging with latent variables and propose a novel loss, the Expected Entity Ratio, to learn models in the presence of systematically missing tags. We show that our approach is both theoretically sound and empirically useful. Experimentally, we find that it meets or exceeds performance of strong and state-of-the-art baselines across a variety of languages, annotation scenarios, and amounts of labeled data. In particular, we find that it significantly outperforms the previous state-of-the-art methods from \citet{mayhew2019named} and \citet{li2021empirical} by $+12.7$ and $+2.3$ F1 score in a challenging setting with only $1,000$ biased annotations, averaged across 7 datasets. We also show that, when combined with our approach, a novel sparse annotation scheme outperforms exhaustive annotation for modest annotation budgets.\footnote{We have published for our implementation and experimental results at \url{https://github.com/teffland/ner-expected-entity-ratio}.}
\end{abstract}

\section{Introduction}

Named entity recognition (NER) is a critical subtask of many domain-specific natural language understanding tasks in NLP, such as information extraction, entity linking, semantic parsing, and question answering.  For large, exhaustively annotated benchmark datasets, this problem has been largely solved by fine-tuning of high-capacity pretrained sentence encoders from massive-scale language modeling tasks \citep{peters2018deep,devlin2018bert,liu2019roberta}. However, fully annotated datasets themselves are expensive to obtain at scale, creating a barrier to rapid development of models in low-resource situations. 

\begin{figure}[t]
  \includegraphics[width=0.48\textwidth]{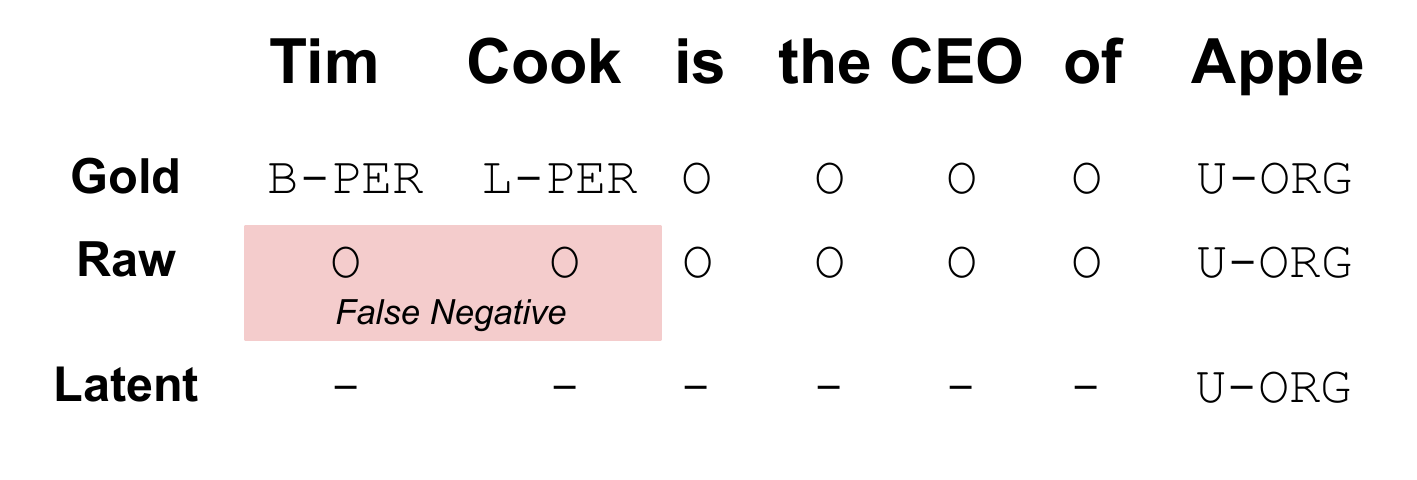}
  \vspace{-25pt}
  \caption{An example low-recall sentence with two entities (one is missing) and its NER tags.
  The Gold row shows the true tags, the Raw row shows a false negative induced by the standard ``tokens without entity annotations are non-entities'' assumption, and the Latent row reflects our view of unannotated tags as latent variables. }

  \label{fig:example}
  \vspace{-15pt}
\end{figure}

Partial annotations, instead, may be much cheaper to obtain. For example, when building a dataset for a new entity extraction task, a domain expert may be able to annotate entity spans with high precision at a lower recall by scanning through documents inexhaustively, creating a higher diversity of contexts and surface forms by limiting the amount of time spent on individual documents. In another scenario studied by \citet{mayhew2019named}, non-speaker annotators for low-resource languages may only be able to recognize some of the more common entities in the target language, but will miss many less common ones. In both of these situations, we wish to leverage partially annotated training data with high precision but low recall for entity spans. Because of the low recall, unannotated tokens are ambiguous and it is not reasonable to assume they are non-entities (the {\tt O} tag).  We give an example of this in Figure~\ref{fig:example}.

We address the problem of training NER taggers with partially labeled, low-recall data by treating unannotated tags as latent variables for a discriminative tagging model.  We propose to combine marginal tag likelihood training~\citep{tsuboi2008training} with a novel discriminative criterion, the Expected Entity Ratio (EER), to control the relative proportion of entity tags in the sentence. The proposed loss is (1) flexibly able to incorporate prior knowledge about expected entity rates under uncertainty; (2) theoretically recovers the true tagging distribution under mild conditions; and (3) easy to implement, fast to compute, and amenable to standard gradient-based optimization. We evaluate our method across 7 corpora in 6 languages along two diverse low-recall annotation scenarios, one of which we introduce. We show that our method performs as well or better than the previous state-of-the-art methods from \citet{mayhew2019named} and the recent work of \citet{li2021empirical} across the studied languages, scenarios, and amounts of labeled entities. Further, we show that our novel partial annotation scheme, when combined with our method, outperforms exhaustive annotation for modest annotation budgets.

\section{Related Works}

A common paradigm for low-recall NER is automatically creating silver-labeled data using outside resources. \citet{bellare2007learning} approach the problem by distantly supervising spans using a database with marginal tag training. \citet{carlson2009learning} similarly use a gazetteer and adapt the structured perceptron \citep{collins2002discriminative} to handle partially labeled sequences, while \citet{nothman2008transforming} use Wikipedia and label-propagation heuristics. \citet{peng2019distantly} also use distant supervision to silver-labeled entities, but use PU-learning with specified class priors to estimate individual classifiers with ad-hoc decoding. \citet{yang2018distantly,Nooralahzadeh2019ReinforcementbasedDO} optimize the marginal likelihood \citep{tsuboi2008training}  of the distantly annotated tags but require gazatteers and some fully labeled data to handle proper prediction of the \texttt{O} tag.  \citet{greenberg2018marginal} use a marginal likelihood objective to pool overlapping NER tasks and datasets, but must exploit cross-dataset constraints. Snorkel~\citep{ratner2020snorkel} uses many sources of weak supervision, but relies on high-recall and overlap to work. In contrast to these works, we do not use outside resources.

Our problem setting has connections to PU-learning, which is classically an approach to classification~\citep{liu2002partially,liu2003building,elkan2008learning,grave2014weakly}, but here we work with tagging structures. Our approach is also related to constraint-satisfaction methods for shaping the model distribution such as CoDL~\citep{chang2007guiding}, used by \citet{mayhew2019named}, and is also related to Posterior Regularization~\citep{ganchev2010posterior}, with main differences being that we do not use the KL-divergence and use gradient-based updates to a nonlinear model instead of closed-form updates to a log-linear model. 

The problem setup from \citet{jie2019better,mayhew2019named} is the same as ours, but \citet{jie2019better} use a cross-validated self-training approach and \citet{mayhew2019named} use an iterative constraint-driven self-training approach to down-weigh possible false-negative \texttt{O} tags, which they show to outperform \citet{jie2019better}. \citet{mayhew2019named} is the current state of the art on the CoNLL 2003 NER datasets~\citep{conll2003} and we compare to their work in the experiments. Recently, \citet{li2021empirical} have published a span-based method that uses negative sampling of non-entity spans, but they do not provide any supporting theoretical guarantees. We also compare to them in the experiments.

\section{Methods}

In this section, we describe the proposed approach. We begin with a description of the problem and notation in \S~\ref{ss:setup}, followed by the NER tagging model in \S~\ref{ss:tagger}. We then describe the supervised marginal tag loss and our proposed auxiliary loss, used for learning on positive-only annotations, in \S~\ref{ss:mmle} and \S~\ref{ss:er}, respectively.  Finally, in \S~\ref{ss:obj} we describe the full objective and give theory showing that our approach recovers the true tagging distribution in the large-sample limit.

\subsection{Problem Setup and Notation}
\label{ss:setup}

We formulate NER as a tagging problem, as is extremely common~\citep{mccallum2003early,lample2016neural,devlin2018bert,mayhew2019named}, inter alia. 
In fully supervised tagging for NER, we are given an input sentence $x_{1:n} = x_1\dots x_n, x_i \in \X$ of length $n$ tokens paired with a sequence $y_{1:n}, y_i \in \Y$ of tags that encode the typed entity spans in the sentence. Following previous work, we use the BILUO scheme \citep{ratinov2009design}. Under this formulation, a NER dataset of fully annotated sentences is a set of pairs of token and tag sequences: $$\Ds^m = \{ (x^{k}_{1:n_k}, y^{k}_{1:n_k}) \}_{k=1}^{m}$$

\subsubsection{Partial Annotations}
\label{sss:partial}

Normally, fully annotated tag sequences are derived from exhaustive annotation schemes, where annotators mark all positive entity spans in the text and then the filler {\tt O} tag can be perfectly inferred at all unannotated tokens. Training a model on such fully annotated data is easy enough with traditional maximum likelihood estimation~\citep{mccallum2003early,lample2016neural}.

In many cases, however, it is desirable to be able to learn on incomplete, partially annotated training data that has high precision for entity spans, but low recall (\S\ref{s:scenarios} discusses two such scenarios). Because of the low recall, unannotated tokens are ambiguous and it is not reasonable to assume they are non-entities (the {\tt O} tag).  Even in this low-recall situation, prior works \citep{jie2019better,mayhew2019named} assume that unannotated tokens are given this non-entity tag. Their approaches then try to estimate which of these tags are ``incorrect'' through self-training-like schemes, iteratively down-weighing the contribution of these noisy tags to the loss with a meta training loop. 

In contrast to prior work, we make no direct assumptions about unannotated tokens and treat all such positions as latent tags. In this view, a partially annotated sentence is a token sequence $x_{1:n}$ paired with a set of observed (tag,position) pairs. Given a sentence $x_{1:n}$, we define
$$y_{\O} \subset \{ (y, i)\;|\;y\in \Y,\; 1 \leq i \leq n \}$$
as the set of observed tags $y$ at positions $i$. For example, in Figure~\ref{fig:example} we would have $y_\O  = \{ (\text{\tt{U-ORG}}, 7) \}$. Under this formulation, we will be given a partially observed dataset:
$$\D^m = \{ (x^{k}_{1:n_k}, y^{k}_{\O_k}) \}_{k=1}^{m}$$
We use data of this form for the rest of the work. 

\subsection{Tagging Model}
\label{ss:tagger}

We use a simple, relatively off-the-shelf tagging model for $p(y_{1:n}|x_{1:n};\theta)$. Our model, BERT-CRF, first encodes the token sequence using a contextual Transformer-based \citep{vaswani2017attention} encoder, initialized from a pretrained language-model objective \citep{devlin2018bert,liu2019roberta}. Given the output representations from the last layer of the encoder, we then score each tag individually with a linear layer, as in \citet{devlin2018bert}. Finally, we model the distribution $p(y_{1:n}|x_{1:n})$ with a linear-chain CRF \citep{lafferty2001conditional}, using the individual tag scores and learned transition parameters $T$ as potentials.  Mathematically, our tagging model is given by:

\begin{gather*}
    h_{1:n} = \text{BERT}(x_{1:n};\theta_{\text{BERT}})\\
    \phi(i,y) = v_{y}^\top h_i\\
    \phi(i,y,y') = \phi(i,y) + T_{y,y'}\\
    p(y|x) = \frac{ \exp \{\sum_{i=1}^{n-1} \phi(i,y_i,y_{i+1}) + \phi(n,y_n) \} } {Z(\phi)}\\
    Z(\phi) = \sum\limits_{\substack{y'_{1:n}\\ \in \Y^n}} \exp \{\sum_{i=1}^{n-1} \phi(i,y'_i,y'_{i+1}) + \phi(n,y'_n) \}
\end{gather*}
where $\phi \in \R^{n \times |\Y| \times |\Y|}$ is the tensor of individual potentials
and $\theta = \{ \theta_{\text{BERT}}, T\} \cup \{ v_y \}_{y \in \Y}$ are the full set of model parameters.

A few important things to note: (1) while we call the encoder ``BERT'', in practice we utilize various BERT-like pretrained transformer language models from the HuggingFace Transformers \citep{huggingface2019gpt2} library; (2) we apply grammaticality constraints to the transition parameters $T$ that cause the model to put zero mass on invalid transitions; and (3) we do not use special start and end states, as pretrained transformers already bookend the  sentence with {\tt SOS} and {\tt EOS} tokens that can be assumed to always be {\tt O} tags. This combined with the transition constraints guarantees that the tagger outputs valid sequences.

We choose this model architecture because it closely reflects recent standard practice in applied NER~\citep{devlin2018bert,huggingface2019gpt2}, where a pretrained transformer is fine-tuned to the tagging dataset. However, we improve this practice by using a CRF layer on top instead of predicting all tags independently. We stress that the additional CRF layer has multiple benefits -- the transition parameters and global normalization improve model capacity and, importantly, prevent invalid predictions. In preliminary experiments, we found that invalid predictions were common in some of the few-annotation scenarios we study here.

\subsection{Supervised Marginal Tag Loss}
\label{ss:mmle}
\newcommand{\Lp}{L_p^{m}}

We train our tagger on partially annotated data by maximizing the marginal likelihood \citep{tsuboi2008training} of the observed tags under the model:
\begin{equation}
    \label{eqn:sup_loss}
    L_p(\theta;\D^m) = \frac{1}{m}\sum_{\substack{(x_{1:n_k}^k, y^k_{\O_k})\\ \in \D^m}} - \log p(y^k_{\O_k}|x^k_{1:n_k};\theta)
\end{equation}

with
\begin{equation}
    \label{eqn:mmle}
    \log p(y_\O|x_{1:n}) = \log \sum\limits_{y_{1:n} \models y_\O} p(y_{1:n}|x_{1:n}) 
\end{equation}
where $y_{1:n} \models y_\O$ means all taggings satisfying the observations $y_\O$. For tree-shaped CRFs, this loss is tractable with dynamic programming.

While it is possible to optimize only this loss for the given partially annotated data, doing so alone has deleterious effects in our scenario -- the resulting model will not learn to meaningfully predict the \texttt{O} tag, by far the most common tag~\citep{jie2019better} and thus fail to have acceptable performance, with high recall at nearly zero precision. We need another term in the loss to encourage the model to predict {\tt O} tags, which we introduce next.

\subsection{Expected Entity Ratio Loss}
\label{ss:er}
\newcommand{\Lu}{L_u^{m}}
\newcommand{\1}{\mathbbm{1}}
\newcommand{\rhohat}{\hat{\rho}_\theta}

As has been observed in prior work~\citep{augenstein2017generalisation,peng2019distantly,mayhew2019named}, the number of named entity tags (versus {\tt O} tags) over the entire distribution of sentences occur at relatively stable rates for different named entity datasets with the same task specification. For any specific dataset, we call this proportion the ``expected entity ratio'' (EER), which  is simply the marginal distribution of some tag $y$ being part of an entity span, $p(y \neq \text{\tt{O}})$. Given an estimate of this EER, $\rho = p(y \neq \text{\tt{O}})$, for the dataset in question, we propose to impose a second loss that directly encourages the tag marginals under the model to match the given EER, up to a margin of uncertainty $\gamma$.
This loss is given by:
\begin{equation}
    \label{eqn:er_loss}
    L_u(\theta;\D^m,\rho,\gamma) = \max \{0, |\rho - \rhohat | - \gamma \}
\end{equation}

\noindent where
\begin{equation}
    \rhohat = \frac{\sum\limits_{\substack{(x_{1:n_k}^k, y^k_{\O_k})\\ \in \D^m}} \E_{p(y^k_{1:n_k}|x_{1:n_k}^k;\theta)}[\sum\limits_{i=1}^{n_k} \1\{y^k_i \neq \text{\tt{O}}\}]}{\sum\limits_{(x_{1:n_k}^k, y^k_{\O_k}) \in \D^m} n_k}
\end{equation}

\noindent is the model's expected rate of entity tags.

For linear-chain CRFs,  the inner expected count \begin{gather*}
    \E_{p(y_{1:n}|x)}[\sum\limits_{i=1}^{n} \1\{y_i \neq \text{\tt{O}}\}]
    = \sum\limits_{i=1}^n\sum\limits_{y \in \Y\setminus\{\text{\tt{O}}\}} p(y_i|x)
\end{gather*}
can be computed exactly, because it factors over the model potentials and reduces to a simple sum over the tag marginals under the model,\footnote{This follows from linearity of expectations: $\E_{y_{1:n}}[\sum_i f(y_i)] = \sum_i \E_{y_{1:n}}[f(y_i)] = \sum_i \E_{y_i}[f(y_i)]$.} and is differentiable.
The outer expectation is not feasible for large datasets on modern hardware, so we approximate it with Monte-Carlo estimates from mini-batches and optimize using stochastic gradient descent~\citep{robbins1951stochastic}.

We also note that the loss in Eqn~\ref{eqn:er_loss} takes the same form as the $\epsilon$-insenstive hinge loss for support vector regression machines~\citep{vapnik1995Slt,Drucker1996SupportVR}, though our use-case is quite different. Additionally, this loss function is differentiable everywhere except at the $\rho \pm \gamma$ points.

\subsection{Combined Objective and Consistency}
\label{ss:obj}

The final loss, presented in Eqn.~\ref{eqn:full_loss}, combines Eqns.~\ref{eqn:sup_loss} and \ref{eqn:er_loss} with a balancing coefficient $\lambda_u$. 

\begin{equation}
    \label{eqn:full_loss}
    L(\theta;\D,\lambda_u,\rho,\gamma) = L_p(\theta;\D) + \lambda_u L_u(\theta;\D,\rho,\gamma) 
\end{equation}

This loss has an intuitive explanation. The supervised loss $L_p$ optimizes the entity recall of the model. The addition of the EER loss $L_u$ further controls the precision of the model. Together, they form a principled objective whose optimum recovers the true distribution under mild conditions.

We now present a theorem 
that gives insight into why the loss in Eqn.~\ref{eqn:full_loss} is justified.
First, we introduce the following set of assumptions:\footnote{We make use of the following definition:
For any finite set ${\cal A}$, define ${\cal A}^+$ to be the set of finite length sequences of symbols drawn from ${\cal A}$. That is, ${\cal A}^+ = \{a_{1:n}: n>0, \; \forall i, a_i \in {\cal A}\}$.}

\begin{assumption}[]
\label{assumption:main}
Assume there are  finite vocabularies of words ${\cal X}$ and tags ${\cal Y}$, and that ${\cal Y}$ contains a special tag {\tt O}. We have some model $p(y_{1:n} | x_{1:n}; \theta)$ with parameter space $\Theta$.
Assume some distribution $p_{X,Y}(x_{1:n}, y_{1:n})$ over sequence pairs $x_{1:n} \in {\cal X}^+, y_{1:n} \in {\cal Y}^+$, and define ${\cal S} = \{ x_{1:n} \in {\cal X}^+ : p_X(x_{1:n}) > 0\}$.
Assume in addition the following:
\begin{description}
\item{(a)} $p_{Y|X}$ is deterministic: that is, for any $x_{1:n} \in {\cal S}$, there exists some $y_{1:n} \in {\cal Y}^+$ such that $p_{Y|X}(y_{1:n} | x_{1:n}) = 1$.
\item{(b)} There is some parameter setting $\theta \in \Theta$ such that $p(y_{1:n} | x_{1:n}; \theta) = p_{Y|X}(y_{1:n} | x_{1:n})$ for all $(x_{1:n}, y_{1:n}) \in {\cal S} \times {\cal Y}^+$.
\item{(c)} We have a set of training examples ${\cal D}^m = \{(x^k_{1:n_k}, y^k_{1:n_k})\}_{k=1}^m$ drawn from the distribution $p_{X}(x_{1:n}) \times \tilde{p}_{Y|X}(y_{1:n} | x_{1:n})$ 
where $\tilde{p}_{Y|X}$
 has the following properties:

{(c1)} No false positives: for all $x_{1:n} \in {\cal S}$, for all $i \in \{1 \ldots n\}$, if $p_{Y|X}(y_i = \text{\tt O} | x_{1:n}) = 1$, then $\tilde{p}(y_i = \text{\tt O} | x_{1:n}) = 1$.
    
{(c2)} Positive entity support: for all $x_{1:n} \in {\cal S}$, for all $i \in \{1 \ldots n\}$, if there is some $y \in {\cal Y}$ such that $y \neq \text{\tt O}$ and $p_{Y|X}(y_i = y | x_{1:n}) = 1$, then $\tilde{p}(y_i = y | x_{1:n}) > 0$, and $\tilde{p}(y_i = \text{\tt O} | x_{1:n}) = 1 - \tilde{p}(y_i = y | x_{1:n})$. That is, only $y$ and {\tt O} are possible under $\tilde{p}$, and the tag $y$ has probability strictly greater than zero.

\end{description}
\end{assumption}

Given these assumptions, define $L^{\infty}$ to be the expected loss under the distribution $\tilde{p}$:
\[
L^{\infty}(\theta; \lambda_u, \rho, \gamma) = \E_{{\cal D}^m \sim \tilde{p}}  \left [ L(\theta; {\cal D}^m, \lambda_u, \rho, \gamma) \right ]
\]

We can then state the following theorem.

\begin{theorem}
\label{thm:main}
Assume that all conditions in assumption~\ref{assumption:main} hold.
Define $\rho = \rho^*$ where $\rho^*$ is the known marginal entity tag distribution, $\gamma = 0$, and $\lambda_u > 0$. Then
for any $\theta \in \arg\min L^{\infty}(\theta; \lambda_u, \rho, \gamma)$, the following holds:
\begin{eqnarray*}
&&\forall (x_{1:n}, y_{1:n}) \in {\cal S} \times {\cal Y}^+, \\
\;\;
&&p(y_{1:n} | x_{1:n}; \theta) = p_{Y|X}(y_{1:n} | x_{1:n})
\end{eqnarray*}
\end{theorem}
The proof of the theorem is in the Appendix.

Intuitively, this result is important because it shows that in the limit of infinite data, parameter estimates optimizing the loss function will recover the correct underlying distribution $p_{Y|X}$. More formally, this theorem is the first critical step in proving consistency of an estimation method based on optimization of the loss function. In particular (see for example Section 4 of \citet{ma2018noise}) it should be relatively straightforward to derive a result of the form
\[
P \left (\lim_{m \rightarrow \infty} d( \hat{p}^m_{Y|X}, p_{Y|X}) = 0  \right ) = 1
\]
under some appropriate definition of distance between distributions $d$, where $\hat{p}^m_{Y|X}$ is the distribution under parameters $\theta^m$ derived from a random sample ${\cal D}^m$ of size $m$. However, for reasons of space we leave this to future work.\footnote{One additional remark: Assumption~\ref{assumption:main} conditions (a) and (b) do not strictly speaking include log-linear models, as probabilities in these models cannot be strictly equal to $1$ or $0$. However, probabilities under these models can approach arbitrarily close to $1$ or $0$; for simplicity we present this version of the theorem here, but a more complete analysis could use techniques similar to those in \citet{della1997inducing} that make use of the closure of the set of distributions of the model, which include points on the boundary.}

\section{Benchmark Experiments}

We evaluate our approach on 7 datasets in 6 languages for two diverse annotation scenarios (14 datasets in total) and compare to strong and state-of-the-art baselines.

\subsection{Corpora}

Our original datasets come from two benchmark NER corpora in 6 languages. We use the English (eng-c), Spanish (esp), German (deu) and Dutch (ned) languages from the CoNLL 2003 shared tasks~\citep{conll2003}. We also use the NER annotations for English (eng-o), Mandarin Chinese (chi), and Arabic (ara) from the Ontonotes5 corpus~\citep{hovy06ontonotes5}. 
    
By studying across this wide array of corpora, we test the approaches in a variety of language settings, as well as dataset and task sizes. The CoNLL corpus specifies 4 entity classes while the Ontonotes corpus has 18 different classes and they span 7.4K to 82K training sentences. 
We use standard train/dev/test document splits. For each corpus, we generate two partially annotated datasets according to the scenarios from \S~\ref{s:scenarios}.

\subsection{Simulated Annotation Scenarios}
\label{s:scenarios}

We simulate two partial annotation scenarios that model diverse real-world situations.  The first is the ``Non-Native Speaker'' (NNS) scenario from \citet{mayhew2019named} and the second, ``Exploratory Expert'' (EE), is a novel scenario inspired by industry. We choose these two samplers to make our results more applicable to practitioners. The simpler alternative -- dropping entity annotations uniformly at random (as in \citet{jie2019better,li2021empirical}) -- is not realistic, leaving an overly diverse set of surface mentions with none of the biases incurred by real-world partial labeling.  While there are other partial annotation scenarios compatible with our method that we could have considered here as well, such as using Wikipedia or gazatteers for silver-labeled supervision, we chose to work with simulated scenarios that allow us to study a large array of datasets without introducing the confounding effects of choices for outside resources.

\subsubsection{Scenario 1: Non-Native Speaker (NNS)}

Our first low-recall scenario is the one proposed by \citet{mayhew2019named}, wherein they study NER  datasets that simulate non-native speaker annotators. To simulate data for this scenario, \citet{mayhew2019named} downsample annotations grouped by mention until a recall of $50\%$.
For example, if ``New York'' is sampled, then all annotations with ``New York'' as their mention in the text are dropped. After the recall is dropped to $50\%$, the precision is lowered to $90\%$ by adding short randomly typed false-positive spans. The reasoning for this slightly more complicated scheme is that it better reflects the biases incurred via non-native speaker annotation.  When non-native speakers exhaustively annotate for NER, they often systematically miss unrecognized entities and occasionally incorrectly annotate false-positive spans.\footnote{It is worth noting that the NNS scenario is also quite close to a silver-labeled scenario using a seed dictionary with 50\% recall, only it has some additional false positive noise.}

The original sampling code used in \citet{mayhew2019named} is not available and we have introduced datasets that were not in their study, so we reimplemented their sampler and used our version across all of our corpora for consistency. We do, however, run their model code on our datasets,
 so our results with respect to their approach still hold.

\subsubsection{Scenario 2: Exploratory Expert (EE)}

In addition to \citet{mayhew2019named}'s non-native speaker scenario, we introduce a signficantly different scenario that reflects another common real-world low-recall NER situation. Though it has not been studied before in the literature, it is inspired by accounts of partially annotated datasets encountered in industry.

In the ``Exploratory Expert'' (EE) scenario, we suppose a new NER task to be annotated by a domain expert with limited time. Here, in the initial ``exploratory'' phase of annotation, the expert may wish to cover more ground by inexhaustively scanning through documents in the corpus, annotating the first few entities they see in a document before moving on, stopping once they have added $M$ total entity spans.  The advantage of this approach is that, by being inexhaustive, the resulting set of mentions and contexts will have more diversity than by using exhaustive annotation. Compared to exhaustive annotation, the disadvantage is annotators may miss entities and the annotations are biased toward the top of documents. 

We simulate this scenario by first removing all annotations from the dataset, then adding back entity spans with the following process. First, we select a document at random without replacement, then scan this document left to right, adding back entity spans with probability $0.8$, until $10$ entities have been added, then moving on to the next random document. The process halts when $M=1,000$ total entity spans have been added back to the dataset.  We note that this assumes that the expert annotators are skimming, sometimes missing entities ($20\%$ of the time), but also assumes that the expert does not make flagrant mistakes and so do not insert random false-positive spans.

An important aspect of this scenario in our experiments is the scale of the number of kept annotations. In previous works~\citep{jie2019better,mayhew2019named,li2021empirical}, the number of kept annotations is not dropped below 50\% of the complete dataset.  By keeping only 1K entities, this scenario is significantly more impoverished than those previously studied (1K entities leaves less than 10\% of annotations for all datasets, ranging from 0.8\% to 8.5\%, depending on the corpus).

\subsection{Approaches}
\label{ss:approaches}

We compare several modeling approaches on the benchmark corpora, detailed below.

\subsubsection{Gold}
For comparison, we report our tagging model trained with supervised sequence likelihood on the original gold datasets. This provides an upperbound on tagging performance and puts any performance degradation from partially-supervised datasets into perspective. We do not expect any of the other methods to outperform this.  

\subsubsection{Raw}
In the {\bf Raw-BERT} baseline, we make the naive assumption that all unobserved tags in the low-recall datasets are the {\tt O}, reflecting the second row of Figure~\ref{fig:example}, and train with supervised likelihood. This is a weak baseline that we expect to have low recall.

\subsubsection{Cost-aware Decoding (Raw+CD)}
This stronger baseline, suggested by a reviewer, explores a simple modification to the Raw baseline at test time: we increase the cost of predicting an {\tt O} tag during inference in an attempt to artificially increase the recall.  That is, we introduce an additional hyperparameter $b_{\text{\tt{O}}} \geq 0$ that is subtracted from the {\tt O} tag potentials, biasing the model away from predicting {\tt O} tags:
\begin{equation*}
    \phi(i,y) = \begin{cases}
        v_{y}^\top h_i - b_{\text{\tt O}} & y = \text{\tt O}\\
        v_{y}^\top h_i & \text{else}
    \end{cases}
\end{equation*}

Intuitively, this approach will work well if the tag potentials consistently rank false negative entity tokens higher than true {\tt O} tokens. To select $b_{\text{\tt{O}}}$, we perform a model-based hyperparameter search~\citep{skopt} using a Gaussian process with 30 evaluations on the validation set F1 score for each dataset's trained Raw-BERT model.

\subsubsection{Constrained Binary Learning (CBL)}

The {\bf CBL} baseline is a state-of-the-art approach to partially supervised NER from \citet{mayhew2019named}. The main idea of the approach is to estimate which {\tt O} tags are false negatives, and remove them from training.

Constrained Binary Learning (CBL) approaches this through a constrained, self-training-like meta-algorithm, based on Constraint-Driven Learning~\citep{chang2007guiding}. The algorithm starts off with a binarized version of the problem ({\tt O} tag vs not) and initializes instance weights of $1$ for all {\tt O} tags.  It then estimates their final weights by iteratively training a model, predicting tags for the training data, then down-weighing some tags based on the confidence of these predictions according to a linear-programming constraint on the total number of allowed {\tt O} tags. At each iteration, the number of allowed {\tt O} tags is decreased slightly, and this loop is repeated until the final target entity ratio (our $\rho$) is satisfied by the weights. A final tagger is then trained on the original tag set using a weighted modification of the supervised tagging likelihood. 

For this method, we used the code exactly as was provided, with the following exception. For all non-english languages, we were not able to obtain the original embeddings used in their experiments, and so we have used language-specific pretrained embeddings from the FastText library~\citep{grave2018learning}.
The base tagging model from \citet{mayhew2019named} utilizes the BiLSTM-CRF approach from \citet{ma2016end}.  The CBL meta-algorithm, however, is agnostic to the underlying scoring architecture of the CRF, and so we test the CBL algorithm both with their BiLSTM scoring architecture and with our BERT-based scoring architecture, which we call {\bf CBL-LSTM} and {\bf CBL-BERT} respectively.  By testing the CBL meta-algorithm with our tagging model, we control for the different modeling choices and get a clear view of how their CBL approach compares to ours.

\subsubsection{Span-based Negative Sampling (SNS)}

The {\bf SNS-BERT} baseline is a recent state-of-the-art approach to partially supervised NER from \citet{li2021empirical}. It uses the same BERT-based encoding architecture, but has a different modeling layer on top. Instead of tagging each token, they instead use a span-based scheme, treating each possible pair of tokens as potential entity and classifying all of the spans independently, using an ad-hoc decoding step based on confidence to eliminate overlapping spans. To deal with the resulting class imbalance ({\tt O} spans are overwhelmingly common) and low-recall entity annotations, they propose to sample spans from the set of unlabeled spans as negatives. While it is possible that they incorrectly sample false negative entities, they argue that this has very low probability. For this method, we used the code as provided but controlled for the same encoding pretrained weights as our other models.

\subsubsection{Expected Entity Ratio (EER)}
The {\bf EER-BERT} model implements our proposed approach, using the proposed tagger (\S~\ref{ss:tagger}) and loss function described in Eqn.~\ref{eqn:full_loss}.

\subsection{Preprocessing}  All datasets came in documents, pre-tokenized into words, with gold sentence boundaries. Recent work~\citep{akbik-etal-2019-pooled,luoma2020exploring} has demonstrated that larger inter-sentential document context is useful for maximizing performance, so we work with full documents instead of individual sentences.\footnote{With the exception of the SNS \citep{li2021empirical} baseline where we had to restrict to sentences because it is $\mathcal{O}(n^2)$ span-based model and could not handle long text sequences, running into memory issues.}
For approaches that used a pretrained transformer, some documents did not fit into the 512 token maximum length. In these cases, we split documents into maximal contiguous chunks at sentence boundaries. Also, for pretrained transformer approaches we expand the tag sequences to match the subword tokenizations.

Because the low-recall data in the EE scenario concentrates annotations at the top of only a few documents, it is possible to identify and omit large unannotated portions of text from the training data. We hypothesize that this will significantly improve model outcomes for the baselines because it significantly cuts down on the number of false negative annotations. Therefore, we explore three preprocessing variants for all EE models: (1) \textbf{all} uses the full dataset as given; (2) \textbf{short} drops all documents with no annotations; and (3) \textbf{shortest} drops all sentences after the last annotation in a document (subsuming \textbf{short}). Model names are suffixed with their preprocessing variants. We note that these approaches do not apply to the NNS scenario, as it has many more annotations spread more evenly throughout the data.

\subsection{Hyperparameters}

All hyperparameters were given reasonable defaults, using recommendations from previous work.  For pretrained transformer models, we used the Huggingface~\citep{huggingface2019gpt2} implementations of {\tt roberta-base} \citep{liu2019roberta} on English datasets and {\tt bert-base-multilingual-cased} \citep{devlin2018bert} for the other languages. The vector representations used by these models are $768$-dimensional and we used matching dimensions for other vector sizes throughout the model. We used a learning rate of $2\times 10^{-5}$ with slanted triangular schedule peaking at 10\% of the iterations~\citep{devlin2018bert}. For batch size, we use the maximum batch size that will allow us to train in memory on a Tesla V100 GPU ($14$ for CoNLL data, $2$ for Ontonote5 data).
We found that training for more epochs than originally recommended~\citep{devlin2018bert} was necessary for convergence and used $20$ epochs for the \textbf{all} variants and $50$ epochs for the significantly smaller \textbf{short} and \textbf{shortest} variants.\footnote{For the CBL-LSTM approach, we use the hyperparameters from \citet{mayhew2019named}: these are more epochs ($45$), and a higher learning rate of $10^{-3}$.}  

The only hyperparameter we adjusted (from a preliminary experiment measuring dev set performance) was setting $\lambda_u = 10$. We originally tried a weight of $\lambda_u = 1$, but then found that the scale of the $L_p$ loss massively overpowered $L_u$, so we increased it to $\lambda_u = 10$, which yielded good performance. We did not try other values after that.

In important contrast to benchmark experiments from prior work~\citep{jie2019better,mayhew2019named}, we do not assume we know the gold entity tag ratio for each dataset when setting $\rho$. Instead, to make the evaluation more realistic, we use a reasonable guess of $\rho = 0.15$ with a margin of uncertainty $\gamma = 0.05$ for all approaches and datasets. We choose this range because it covers most of the gold ratios observed in the datasets.\footnote{In early experiments we found that the CBL code from \citet{mayhew2019named} used the gold ratio {\it plus} $0.05$. This additional $0.05$ turned out to be critical to getting competitive performance, so in practice we use a $\rho=0.2$ for CBL.} 

\subsection{Results}

\begin{table*}[t!]
\begin{center}
\begin{tabular}{llllllll|c}
\toprule

Approach / Language & eng-c & deu & esp & ned & eng-o & chi & ara & avg\\
\midrule
Gold-BERT-all & 92.7 & 83.9 & 88.3 & 91.1 & 90.7 & 79.4 & 72.9 & 85.6\\
Gold-SNS-BERT-all & 91.1 & 82.3 & 87.9 & 89.5 & 89.7 & 77.1 & 62.1 & 82.8\\
\rule{0pt}{1.5\normalbaselineskip}
& \multicolumn{8}{c}{Non-Native Speaker Scenario (NNS): Recall=$50\%$, Precision=$90\%$ } \\
\cmidrule{2-9}
Raw-BERT-all  & 81.9 & 69.1 & 71.2 & 70.1 & 68.0 & 61.9 & 52.8 & 67.9\\
Raw+CD-BERT-all & 86.3 & 78.4 & 79.9 & 77.2 & 80.9 & 64.9 & 60.1 & 75.4\\
CBL-LSTM-all & 79.2 & 38.4 & 54.6 & 48.2 & 67.9 & 53.5 & 39.4 & 54.5\\
CBL-BERT-all & 84.8 & \textbf{77.5} & 78.7 & 75.3 & 76.3 & \textbf{68.9} & \textbf{61.9} & 74.8\\
SNS-BERT-all & 86.0 & 77.0 & 80.8 & \textbf{77.9} & 81.5 & 66.4 & 56.0 & 75.1\\
EER-BERT-all & \textbf{88.0} & 77.3 & \textbf{80.9} & 76.9 & \textbf{84.5} & 66.6 & 56.6 & \textbf{75.8}\\
\rule{0pt}{1.5\normalbaselineskip}
& \multicolumn{8}{c}{Exploratory Expert Scenario (EE): $1,000$ Annotations } \\
\cmidrule{2-9}
Raw-BERT-all      & 0.4  & 02.6 & 00.7 & 0.0  & 0.4 & 2.4 & 5.3 & 1.7\\
Raw-BERT-short    & 44.1 & 37.2 & 44.4 & 0.0  & 28.4 & 32.4 & 15.4 & 28.8\\
Raw-BERT-shortest & 80.7 & 65.4 & 73.0 & 69.1 & 67.5 & 57.1 & 42.0 & 65.0\\
Raw+CD-BERT-shortest & 82.4 & 67.9 & 76.6 & 70.0 & 68.9 & 58.3 & 43.9 & 66.9\\
CBL-LSTM-all      & 60.2 & 27.5 & 41.2 & 33.3 & 23.1 & 29.9 & 15.3 & 32.9\\
CBL-LSTM-shortest & 67.8 & 20.1 & 36.2 & 26.7 & 42.0 & 24.6 & 9.7 & 32.4\\
CBL-BERT-all      & 36.4 & 52.8 & 40.9 & 52.5 & 22.4 & 29.3 & 20.8 & 36.4\\
CBL-BERT-short    & 43.7 & 64.7 &  56.4  & 60.8 & 16.0 & 31.2 & 30.2 & 43.3\\
CBL-BERT-shortest & 80.6 & 65.1 & 74.7 & 71.2 & 28.4 & 53.6 & 39.2 & 59.0\\
SNS-BERT-all & 59.5 & 63.8 &  70.8 &  70.3 & 14.0 &  28.8 & 0.0  & 43.9\\
SNS-BERT-short & 64.4 & 62.6 & 70.8 & 64.1 & 40.7 & 46.4 & 0.0 & 49.9\\
SNS-BERT-shortest & 83.9 & 70.1 & 76.8 & 77.1 & 75.6 & 63.3 & 40.7 & 69.6\\
EER-BERT-all      & 86.3 & 73.2 & \textbf{80.2} & 80.2 & 61.2 & 56.2 & 42.9 & 68.6\\
EER-BERT-short    & \textbf{89.0}$^\dagger$ & 72.2 & 76.5 & \textbf{80.3}$^\dagger$ & \textbf{75.9} & 61.4 & \textbf{46.8}$^\dagger$ & \:\,\textbf{71.7}$^\dagger$\\
EER-BERT-shortest & 87.3$^\dagger$ & \textbf{73.6}$^\dagger$ & 76.5 & 74.2 & 74.0 & \textbf{64.3} & 42.1 & 70.3 \\
\bottomrule
\end{tabular}
\caption{Benchmark test set F1 scores across different languages and annotation scenarios. Best models in bold.\\
$\dagger$ indicates that for EE the test F1 score is statistically signficantly better than SNS-BERT-shortest ($p<0.01$) (details in footnote~\ref{bootstrap}). Other pairs between SNS-BERT-shortest and EER-BERT-short/shortest were not signficant.}
\label{table:benchmark}
\vspace{-10pt}
\end{center}
\end{table*}

The results of our evaluation are presented in Table~\ref{table:benchmark}. The first row shows the result of training our tagger with the original gold data. These results are competitive with previously published results from similar pretrained transformers~\citep{devlin2018bert} that do not use ensembles or NER-specific pretraining~\citep{luoma2020exploring,baevski2019cloze,yamada2020luke}. Interestingly, we also found that our tagging CRF outperformed the span-based independent distribution of \citet{li2021empirical} on all gold datasets.

{\bf NNS Performance.} The second set of rows shows test F1 scores of models from \S~\ref{ss:approaches} for the NNS sampled datasets. We first note that the CBL-LSTM approach from \citet{mayhew2019named} significantly underperformed for all non-english languages (and are much lower than the results from their paper with similar data). We used their code as is, only changing the pretrained word vectors, and so suspect that this is due to lower quality word vectors obtained from FastText instead of their custom-fit vectors.  This is confirmed by the results of using their CBL meta-algorithm with our proposed tagging architecture, which is competitive with EER-BERT in this setting.
Otherwise, we found that all strong baselines and our method performed quite similarly. This suggests that performance in the NNS regime with relatively high recall (50\%) and little label noise per positively labeled mention is not bottlenecked by approaches to resolving missing mentions. Further improvements in this regime will likely come from other sources, such as better pretraining or supplemental corpora. Because of this we recommend that future evaluations for partially supervised NER focus on more impoverished annotation counts, such as the EE scenario we study next.

{\bf EE Performance.} In the third group of rows, we show test F1 scores for each model using the more challenging EE scenario with only $1,000$ kept annotations. In this setting, using the dataset as is for supervised training (Raw-BERT-all), fails to converge, but smarter preprocessing largely alleviates this problem, with Raw-BERT-shortest obtaining an average F1 of $65.0$. Adding cost-aware decoding (Raw+CD-BERT-shortest) further improves upon the standard baseline (F1 $66.9$).  

Even with only $1,000$ biased and incomplete annotations -- less than $10\%$ of the original annotations for all datasets -- we find that our approach (EER-BERT-short) still achieves an F1 score of $71.7$ on average. This outperforms the best strong baselines: Raw+CD, CBL, and SNS, by $4.8$, $12.7$, and $2.3$ F1 score, respectively. The closest baseline, SNS-BERT-shortest from \citet{li2021empirical}, is competitive with EER-BERT-short on four of the datasets, but performs significantly worse on the other three as well as overall,\footnote{\label{bootstrap}We assessed significance between model pairs using a percentile bootstrap of F1 score differences, resampling test set documents with replacement 100K times~\citep{efron1994introduction} and measuring the paired F1 scores differences of EER-BERT-short/shortest and SNS-BERT-shortest. Significance was assessed by whether the two-sided 99\% confidence interval contained 0.0. To assess overall significance, we concatenated the test datasets before bootstrapping.}
leading us to conclude that our method has a performance edge in this regime. 
Further, EER-BERT-short performs only $4.1$ average F1 worse on EE data than EER-BERT-all on NNS data. We also note that EER-BERT-shortest significantly outpeformed SNS-BERT-shortest on two datasets, but failed to reject the null hypothesis overall.

Another important finding is that EER-BERT is much more robust to preprocessing choices than the baselines. The baselines all view missing entities as {\tt O} tags/spans (at least to start) and these relatively common false negatives severely throw off convergence. By removing most of the unannotated text with preprocessing, we effectively create a much smaller corpus that has nearly 80\% recall (for shortest). In contrast, EER-BERT's view of the data makes no assertions about the class of individual unobserved tokens and so is less sensitive to the relative proportion of false negative annotations. This is useful in practice, as our approach should better handle partial annotation scenarios with wider varieties of false negative proportions that may not be so easily addressed with simple preprocessing.

{\bf Speed.} A pragmatic appeal of our approach compared to CBL~\citep{mayhew2019named} is training time. On NNS data, EER-BERT-all is on average $7.6$ times faster than CBL-BERT-all and on EE data EER-BERT-short is $2.2$ times faster than CBL-BERT-shortest, even though it uses more data. This is because EER does not require a costly outer self-training loop.
\footnote{We unfortunately cannot comment on relative speed of SNS because runtimes cannot be inferred from the SNS code output, though we do not expect a fundamental speed advantage of one over the other, as neither use self-training.}

{\bf Conclusion.} These results illustrate that our approach outperforms the previous strong and state-of-the-art baselines in the challenging low-recall EE setting with only 1K annotations while also being more robust to the relative proportions of false negatives in the training corpus.\footnote{We also note that the EE scenario averages for all models are significantly affected by the poor performance on the Arabic  Ontonotes5 (ara) dataset. After further inspection of the training curves, we found that all models exhibited very slow convergence on this dataset and/or failed to converge in the allotted number of epochs.}

\subsection{Analysis of EER hyperparams}

Recall that the definition of our EER loss in Eqn.~\ref{eqn:er_loss} defines an acceptable region $\rhohat \in [\rho-\gamma,\rho+\gamma]$ of learned models and that in our this experiment, we used $\rho = 0.15$ and $\gamma=0.05$ for all datasets, regardless of the true entity ratios $\rho^*$.  Two interesting questions then are (1) ``how sensitive is the procedure to choices of $\rho$ and $\gamma$?''; and (2) ``how closely do the final learned models reflect the true entity ratios for the data?''. We address these next.

\subsubsection{Robustness to choices of $\rho$ and $\gamma$}

To study robustness we varied choices of $\rho$ and $\gamma$ for EER-BERT-short on the CoNLL English EE dataset with three randomly sampled datasets. Table~\ref{table:robustness} shows test F1 scores across seeds for various settings of $\rho \pm \gamma$. We first show three point estimates with $\gamma = 0.0$, the first at $\rho = \rho^* = 0.23$, then shifted around $\rho^*$ left and right to $\rho = 0.15$ and $\rho = 0.30$, respectively. We then widen the range with $\gamma = 0.05$ and show the benchmark result $\rho = 0.15$, followed by shifts of $\rho \pm 0.1$. Finally we show a very wide range of $\rho = 0.15, \gamma = 0.15$. 

\begin{table}[H]
\begin{tabular}{lcccc}
\toprule
\multicolumn{5}{c}{ Varying EER HPs Test F1 Scores} \\
$[\rho-\gamma, \rho+\gamma]$& RS0 & RS1 & RS2 & Avg.\\
\midrule
$[0.23,0.23]^*$ & 86.8 & 87.4 & 87.0 & 87.1\\
$[0.15,0.15]$ & 89.3 & 87.1 & 87.8 & 88.1 \\
$[0.30,0.30]$  & 79.1 & 79.4 & 79.7 & 79.4\\
$[0.10,0.20]^\dagger$ & 87.6 & 88.2 & 87.8 & 87.9\\
$[0.20,0.30]$  & 83.9 & 83.8 & 84.1 & 83.9\\
$[0.00,0.10]$  & 89.2 & 87.1 & 87.8 & 88.0\\
$[0.00,0.30]$  & 83.9 & 83.8 & 84.0 & 83.9\\

\bottomrule
\end{tabular}
\label{table:robustness}
\caption{CoNLL English EE EER-short test set F1 across three randomly sampled datasets. $*$: $\rho = \rho^*$. $\dagger$: benchmark experiment setting.}
\end{table}

From the table we can glean two interesting points. The first is that in settings where the high end of range of acceptable EER's is greater then $\rho^*$ (when $\rho + \gamma = 0.30$) there is a substantial drop in performance (mean = $82.3$). The second is that the complement group of settings, where $\rho + \gamma \leq \rho^*$ are all high-performing with little variance (mean = $87.8$, std = $0.4$). Together they suggest that the true sensitivity of the proposed EER approach to the high end of the interval and that it is best to conservatively estimate that value, whereas the low end of the range is unimportant. This result agrees well with the intuitions provided in \S~\ref{ss:obj}: since $L_p$ is encouraging models with high recall without regard for precision ($\rhohat \rightarrow 1$), it is best to set $\rho + \gamma$ such that $L_u$ introduces a tension in the combined loss by encouraging $\rhohat \leq \rho^*$. This is not the whole story, however, as we discuss next.

\subsubsection{Convergence towards $\rho^*$} 

The results from the previous experiment suggests that $L_u$ simply serves to drive $\rhohat \rightarrow \rho + \gamma$. Since we used $\rho + \gamma = 0.2$ for all datasets in the benchmark, we would then expect to see a result that $\rhohat \approx 0.2$ for all models. 

We tested this hypothesis by calculating the entity ratio $\rhohat$ of final trained EER-BERT-short models for the EE datasets (leaving out ara, since it failed to converge) and calculated the average difference of each $\rhohat$ with respect to the corresponding true $\rho^*$, resulting in mean absolute error of only $0.018$. This is much closer on average than if the models just converged to $0.2$ (the mean absolute error then would be $0.048$), indicating that our approach tends to converge more closely to the true entity ratio $\rho^*$ than the estimate given by $\rho + \gamma$. In particular, we found that all final models had $\rhohat < 0.2$ except CoNLL English, where $\rhohat = 0.23$, quite close to the gold $\rho^*$ even though it was outside of the target range. This result is encouraging in that it suggests the EER loss, in balance with the supervised marginal tag loss, does more to recover $\rho^*$ than just drive $\rhohat \rightarrow \rho + \gamma$.

\section{EE vs. Exhaustive Experiments}

In situations where we only have partially annotated data without the option for exhaustive annotations, the utility of being able to train with the data as provided is self-evident. However, given the potential upsides of partial annotation relative to exhaustive annotation -- mentally less taxing and increased contextual diversity for a fixed annotation budget -- it is natural to ask whether it is actually {\it better} to go with a sparse annotation scheme.

\subsection{Annotation Speed User Study}

We begin with a user study of annotation speed, comparing EE to the standard exhaustive annotation scheme. Following methodology from \citet{li-etal-2020-active-learning}, we recorded 8 annotation sessions from 4 NLP researchers familiar with NER. Using the Ontonotes5 English corpus, we asked each annotator to annotate for two 20 minute sessions using the BRAT~\citep{brat} annotation tool, one exhaustively and the other following the EE scheme. We split documents into two randomized groups and systematically varied which group was annotated with each scheme and in what order to control for document and ordering variation effects. Then, for each annotator, we measured the number of annotated entities per minute for both schemes and report the ratio of EE annotations per minute to exhaustive annotations per minute (i.e., the relative speed of EE to exhaustive). We found that, although speed varied greatly between annotators (ranging from roughly $4$ annotations/min to $9$ annotations/min across sessions), EE annotation and exhaustive annotation were essentially the same speed, with EE being $3\%$ faster on average. Thus we may fairly compare exhaustive and EE schemes using model performance at the same number of annotations, which we do next.\footnote{
The exact number of annotated entities among the four annotation sessions for EE were 91, 90, 109, and 179. For Exhaustive the matching annotation counts were 83, 85, 117, and 170.
}

\subsection{Performance Learning Curves}

In this experiment, we compare the best traditional supervised training from the benchmark (Raw-BERT-shortest) with our proposed approach (EER-BERT-short) on EE-annotated and exhaustively annotated documents from CoNLL'03 English (eng-c) at several annotation budgets, $M \in \{100\ (0.4\%),$ $ 500\ (2.1\%),$ $ 1K\ (4.3\%),$ $ 5K\ (21.3\%),$ $ 10K\ (42.6\%)\}$.
For each annotation budget, we sampled three datasets with different random seeds for both annotation schemes and trained both modeling approaches. This allows us to study how all four combinations of annotation style and training methods perform at varying magnitudes of annotation counts. In addition to low-recall annotations, we compared our EER approach to supervised training on the gold data.

\begin{figure}[b!]
  \vspace{-20pt}
  \includegraphics[width=0.52\textwidth]{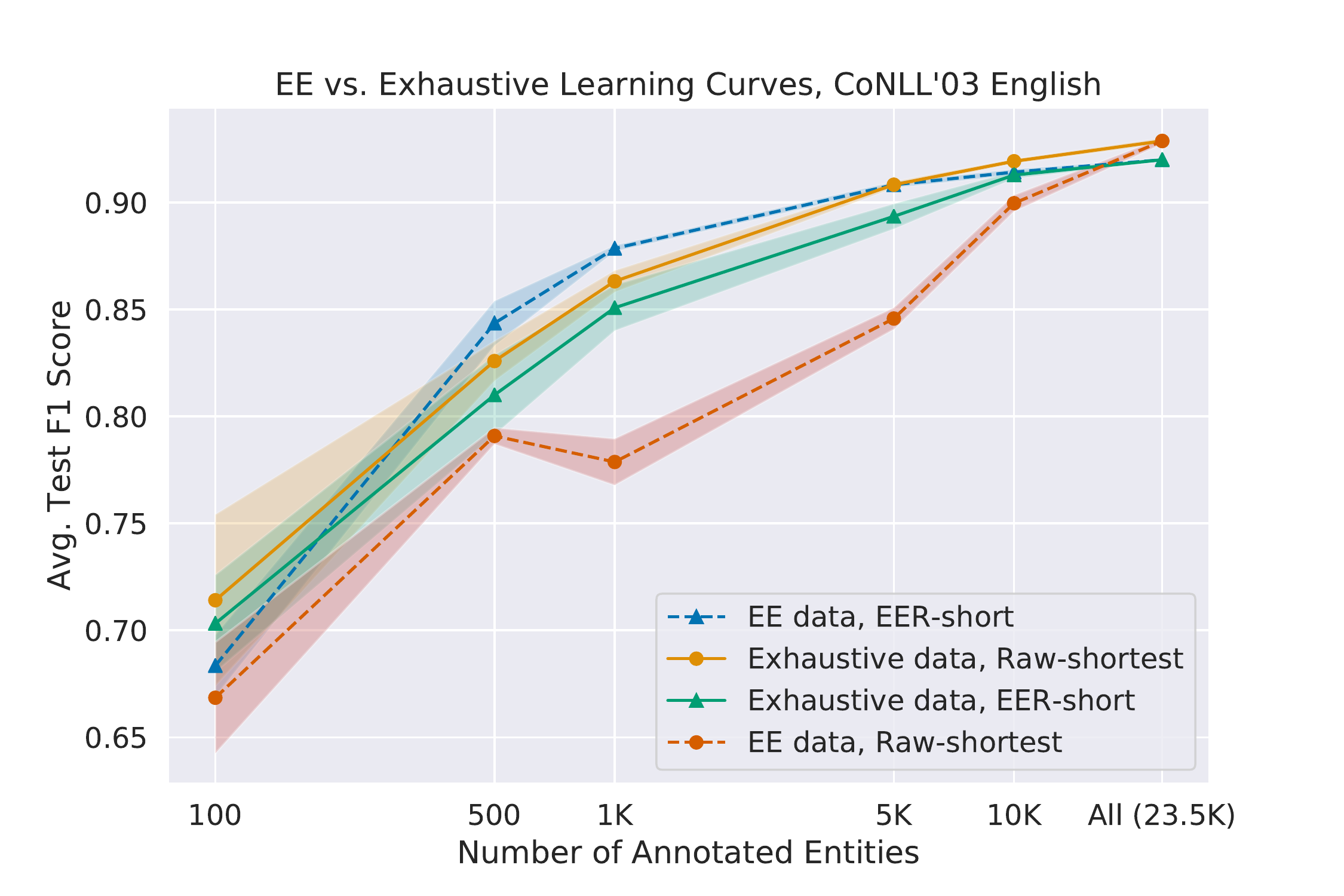}
  \vspace{-20pt}
  \caption{Test performance as a function of the number of observed training annotations for the Exhaustive vs. EE annotation on CoNLL English. Lines are averages and shaded regions are $\pm 1$ standard error.}
  \label{fig:lc}
\end{figure}

In Figure~\ref{fig:lc}, we show learning curves for the average test performance of all four annotation/training variants. From the plot, we can infer several points. First, on EE-annotated data, using our EER loss substantially outperforms traditional likelihood training at all amounts of partial annotation, but the opposite is true on exhaustively annotated data. This indicates that the training method should be tailored to the annotation scheme.

The comparison between EE data with EER training versus exhaustive data with likelihood training is more nuanced. At only 100 annotations, exhaustive annotation worked best on average in our sample, but all methods exhibit high variance due to the large variation in {\it which} entities were annotated. Interestingly, at modest sizes of only 500 and 1K annotations, EE annotated data with our proposed EER-short approach outperformed exhaustive annotation with traditional supervised training, with gains of $+1.8$ and $+1.5$ average F1 for 500 and 1K annotations, respectively. These results, however, reverse as the annotation counts grow: at 5K annotations, the two approaches perform the same ($90.8$) and, at even larger annotation counts, exhaustive annotation with traditional training outperforms our approach by $+0.5$ at 10K annotations and $+0.8$ on the gold dataset. This indicates that EE annotation, paired with our EER loss, is competitive and potentially advantageous to exhaustive annotation and traditional training at modest annotation counts, but that exhaustive annotation with traditional training is better at large annotation counts. This suggests that a hybrid annotation approach where we sparsely annotate data at first, but eventually switch to exhaustive annotations as the process progresses, is a promising direction of future work. We note that our EER loss can easily incorporate observed {\tt O} tags from exhaustively annotated documents in $y_\O$ and so would work in this setup without modification.

\section{Conclusions}

We study learning NER taggers in the presence of partially labeled data and propose a simple, fast, and theoretically principled approach, the Expected Entity Ratio loss, to deal with low-recall annotations. We show empirically that it outperforms the previous state of the art across a variety of languages, annotation scenarios, and amounts of labeled data. Additionally, we give evidence that sparse annotations, when paired with our approach, are a viable alternative to exhaustive annotation for modest annotation budgets. 

Though we study two simulated annotation scenarios to provide controlled experiments, our proposed EER approach is compatible with a variety of other incomplete annotation scenarios, such as incidental annotations (e.g., from web links on Wikipedia), initialized by seed annotations from incomplete distant supervision/gazatteers, or embedded as a learning procedure in an active/iterative learning framework, which we intend to explore in future work. 

\section*{Acknowledgements}

We would like to thank Chris Kedzie, Giannis Karamanolakis, and the reviewers for helpful conversations and feedback.

%%%%%%%%%%%%%%%%%%%%%%%%%%%%%%%%
%%%%%%%%%%%%%%%%%%%%%%%%%%%%%%%%
%%%%%%%%%%%%%%%%%%%%%%%%%%%%%%%%
%%%%%%%%%%%%%%%%%%%%%%%%%%%%%%%%
%%%%%%%%%%%%%%%%%%%%%%%%%%%%%%%%
%%%%%%%%%%%%%%%%%%%%%%%%%%%%%%%%
\bibliography{tacl2018}
\bibliographystyle{acl_natbib}

% \clearpage
\begin{appendix}
\section{Appendix: Proof of Theorem~\ref{thm:main}}
\label{app:proof}

{\em Proof of Theorem~\ref{thm:main}}: We have
\[
L^{\infty}(\theta; \lambda_u, \rho, \gamma)
= g(\theta) + h(\theta) 
\]
where $g(\theta) = \E [L_p(\theta;\D^m)]$ and $h(\theta) = \E [ \lambda_u L_u(\theta;\D^m,\rho,\gamma) ]$. Note that
\begin{equation}
g(\theta) = \sum_{x_{1:n}, y_{1:n}} 
\tilde{p}(x_{1:n}, y_{1:n}) g'(x_{1:n}, y_{1:n}, \theta)
\label{eq:g}
\end{equation}
where $\tilde{p}(x_{1:n}, y_{1:n}) = p_X(x_{1:n}) \times \tilde{p}(y_{1:n} | x_{1:n})$ and
\begin{equation}
g'(x_{1:n}, y_{1:n}, \theta) = 
- \log
\sum_{y'_{1:n}  \models y_{1:n}} p(y'_{1:n} | x_{1:n}; \theta)
\label{eq:gp}
\end{equation}
Define $\theta^*$ to be such that $\forall x_{1:n} \in {\cal X}$, $\forall y_{1:n}$,  $p(y_{1:n} | x_{1:n}; \theta^*) = p_{Y|X}(y_{1:n} | x_{1:n})$ (by assumption~\ref{assumption:main}(b) such a parameter setting must exist).
The following properties are easily verified to hold: (1) $\forall \theta, g(\theta) \geq 0, h(\theta) \geq 0$ and (2) $g(\theta^*) = h(\theta^*) = 0$.  Hence $\theta^*$ is a minimizer of $g(\theta) + h(\theta)$. 

We now show that {\em any} minimizer $\theta'$ of $g(\theta) + h(\theta)$ must satisfy the property that $\forall x_{1:n} \in {\cal X}, \forall y_{1:n}, \; p(y_{1:n} | x_{1:n}; \theta^{'}) = p_{Y|X}(y_{1:n} | x_{1:n})$. For $\theta'$ to be a minimizer of $g(\theta) + h(\theta)$ it must be the case that $g(\theta') = h(\theta') = 0$. We then note the following steps:

(i) By Lemma~\ref{lemma:one}, if $g(\theta') = 0$ it must hold that $
\forall x_{1:n} \in {\cal X}, \forall i \in \{1 \ldots n\}$ such that $p_{Y|X}(y_i = y | x_{1:n}) = 1$ and $y \neq \hbox{\tt o}$, $p(y_i = y | x_{1:n}; \theta') = 1.$

(ii) It remains to be shown that $
\forall x_{1:n} \in {\cal X}, \forall i \in \{1 \ldots n\}$ such that $p_{Y|X}(y_i = y | x_{1:n}) = 1$ and $y = \hbox{\tt o}$, $p(y_i = y | x_{1:n}; \theta') = 1.$

(iii) Property (ii) follows from (i) through proof by contradiction. If $\exists\ x_{1:n} \in {\cal X}$ together with $i \in \{1 \ldots n\}$ such that $p_{Y|X}(y_i = y | x_{1:n}) = 1$ and $y = \hbox{\tt o}$, and $p(y_i = y | x_{1:n}; \theta) < 1$ it must be the case that $h(\theta') > 0$, because the expected number of $\hbox{\tt o}$ tags under $\theta'$ is strictly less than the expected number of $\hbox{\tt o}$ tags under $p_{Y|X}$. $\qed$

\begin{lemma}
\label{lemma:one}
Define $g(\theta)$ and $g'(x_{1:n}, y_{1:n}, \theta)$ as in Eqs.~\ref{eq:g} and~\ref{eq:gp} above.
\commentout{
\[
g(\theta) = \sum_{x_{1:n}, y_{1:n}} 
\tilde{p}(x_{1:n}, y_{1:n}) g'(x_{1:n}, y_{1:n}, \theta)
\]
where $\tilde{p}(x_{1:n}, y_{1:n}) = p_X(x_{1:n}) \times \tilde{p}(y_{1:n} | x_{1:n})$ and
\[
g'(x_{1:n}, y_{1:n}, \theta) = 
- \log
\sum_{y'_{1:n}  \models y_{1:n}} p(y'_{1:n} | x_{1:n}; \theta)
\]}
For any value of $\theta$ such that $g(\theta) = 0$,
$\forall x_{1:n} \in {\cal X}$, $\forall i \in \{1 \ldots n\}$ such that $p_{Y|X}(y_i = y | x_{1:n}) = 1$ and $y \neq \hbox{\tt o}$, $p(y_i = y | x_{1:n}; \theta) = 1.$
\end{lemma}

{\em Proof:} If $g(\theta) = 0$, then for all $x_{1:n}, y_{1:n}$ such that $\tilde{p}(x_{1:n}, y_{1:n}) > 0$, it must be the case that $- \log
\sum_{y'_{1:n}  \models y_{1:n}} p(y'_{1:n} | x_{1:n}; \theta)
= 0$ and hence $\sum_{y'_{1:n}  \models y_{1:n}} p(y'_{1:n} | x_{1:n}; \theta)
= 1$. The proof is then by contradiction: if there exists some $x_{1:n} \in {\cal X}, i \in \{1 \ldots n\}$ such that $p_{Y|X}(y_i = y | x_{1:n}) = 1$ and $y \neq \hbox{\tt o}$ and $p(y_i = y | x_{1:n}; \theta) < 1$, it must be the case that there exists some $y_{1:n}$ such that $\tilde{p}(x_{1:n}, y_{1:n}) > 0$, $y_i = y$, and $\sum_{y'_{1:n}  \models y_{1:n}} p(y'_{1:n} | x_{1:n}; \theta)
< 1$.
\end{appendix}

\end{document}